***Title page***

***Title***

A new method of modeling the multi-stage decision-making process of CRT using machine learning with uncertainty quantification

***Running Head***

Multi-stage ML to predict CRT response

***Authors***

Kristoffer Larsen[a], B.S.

kalarsen@mtu.edu

Chen Zhao[b], Ph.D.

czhao4@kennesaw.edu

Joyce Keyak[c], Ph.D.

jhkeyak@hs.uci.edu

Qiuying Sha[a], Ph.D.

qsha@mtu.edu

Diana Paez[d], MsEd., M.D.

d.paez@iaea.org

Xinwei Zhang[e], M.D., Ph.D.

zhangxinwei_njmu@sina.com

Guang-Uei Hung[f], M.D.

106143@gmail.com

Jiangang Zou[e], M.D., Ph.D., FHRS

jgzou@njmu.edu.cn

Amalia Peix[g], Ph.D., M.D.

atpeix@gmail.com

Weihua Zhou[h,i], Ph.D.

whzhou@mtu.edu

***Institutions***

[a] Department of Mathematical Sciences, Michigan Technological University, Houghton, MI, USA

[b] Department of Computer Science, Kennesaw State University, Marietta, GA, USA

[c] Department of Radiological Sciences, Department of Biomedical Engineering, and Department of Mechanical and Aerospace Engineering, University of California, Irvine, CA, USA

[d] Nuclear Medicine and Diagnostic Imaging Section, Division of Human Health, Department of Nuclear Sciences and Applications, International Atomic Energy Agency, Vienna, Austria

[e] Department of Cardiology, The First Affiliated Hospital of Nanjing Medical University, Nanjing, Jiangsu, China

[f] Department of Nuclear Medicine, Chang Bing Show Chwan Memorial Hospital, Changhua, Taiwan

[g] Nuclear Medicine Department, Institute of Cardiology, La Habana, Cuba

[h] Department of Applied Computing, Michigan Technological University, Houghton, MI, USA

[i] Center for Biocomputing and Digital Health, Institute of Computing and Cybersystems, and Health Research Institute, Michigan Technological University, Houghton, MI, USA

***Address for correspondence***

Weihua Zhou

E-mail: whzhou@mtu.edu

Address: 1400 Townsend Drive, Houghton, MI 49931, USA

Or

Amalia Peix

E-mail: atpeix@gmail.com

Address: Nuclear Medicine Department, Institute of Cardiology, 17 No. 702, Vedado, CP 10 400, La Habana, Cuba

Or

Jiangang Zou

E-mail: jgzou@njmu.edu.cn

Address: Guangzhou Road 300, Nanjing, Jiangsu, China 210029

***First Author***

Kristoffer Larsen (Student)

E-mail: kalarsen@mtu.edu

Address: 1400 Townsend Drive, Houghton, MI 49931, USA

*Abstract Page*

## ABSTRACT

*Aims*. Current machine learning-based (ML) models usually attempt to utilize all available patient data to predict patient outcomes while ignoring the associated cost and time for data acquisition. The purpose of this study is to create a multi-stage machine learning model to predict cardiac resynchronization therapy (CRT) response for heart failure (HF) patients. This model exploits uncertainty quantification to recommend additional collection of single-photon emission computed tomography myocardial perfusion imaging (SPECT MPI) variables if baseline clinical variables and features from electrocardiogram (ECG) are not sufficient.

*Methods*. 218 patients who underwent rest-gated SPECT MPI were enrolled in this study. CRT response was defined as an increase in left ventricular ejection fraction (LVEF) > 5% at a $6\pm1$ month follow-up. A multi-stage ML model was created by combining two ensemble models: Ensemble 1 was trained with clinical variables and ECG; Ensemble 2 included Ensemble 1 plus SPECT MPI features. Uncertainty quantification from Ensemble 1 allowed for multi-stage decision-making to determine if the acquisition of SPECT data for a patient is necessary. The performance of the multi-stage model was compared with that of Ensemble models 1 and 2.

*Results*. The response rate for CRT was 55.5% (n = 121) with overall male gender 61.0% (n = 133), an average age of 62.0±11.8, and LVEF of 27.7±11.0. The multi-stage model performed similarly to Ensemble 2 (which utilized the additional SPECT data) with AUC of 0.75 vs. 0.77, accuracy of 0.71 vs. 0.69, sensitivity of 0.70 vs. 0.72, and specificity 0.72 vs. 0.65, respectively. However, the multi-stage model only required SPECT MPI data for 52.7% of the patients across all folds.

*Conclusions*. By using rule-based logic stemming from uncertainty quantification, the multi-stage model was able to reduce the need for additional SPECT MPI data acquisition without sacrificing performance.

**ABBREVIATIONS**

| | |
|---|---|
| CRT | Cardiac resynchronization therapy |
| ECG | Electrocardiogram |
| GDMT | Guideline-directed medical therapy |
| HF | Heart failure |
| LBBB | Left bundle branch block |
| LVEF | Left ventricular ejection fraction |
| LVMD | Left ventricular mechanical dyssynchrony |
| MPI | Myocardial perfusion imaging |
| SPECT | Single-photon emission computed tomography |
| ML | Machine Learning |

## INTRODUCTION

Approximately 6.2 million adults in the USA have heart failure (HF)[1], with an estimated treatment cost of $43.6 billion in 2020[2]. Cardiac resynchronization therapy (CRT) is a standard treatment for HF that coordinates the functions of the left and right ventricles at an average cost of > $60,000 per patient[3]. In the USA alone, the number of CRT procedures is >60,000 per year[3]. Despite the prevalence of CRT, a significant proportion (30-40%) of patients who receive CRT do not respond to the treatment[4] because the criteria for selecting patients for treatment have not been fully optimized among other factors, such as the complexity of HF. Guidelines recommend CRT for patients with left ventricular ejection fraction (LVEF) ≤ 35%, sinus rhythm, left bundle-branch block (LBBB) with QRS duration ≥ 150 ms, and New York Heart Association (NYHA) class II, III, or ambulatory IV symptoms on guideline-directed medical therapy (GDMT)[5]. Selection of patients based on electrical dyssynchrony, such as measured from electrocardiogram (ECG), is not sufficient; mechanical dyssynchrony, such as measured from SPECT MPI has shown importance for CRT despite not being used in current guidelines[6,7]. However, SPECT MPI has associated costs and radiation exposure. Therefore, CRT patient selection can be improved by utilizing personalized prediction methods which can incorporate SPECT MPI.

Machine learning (ML) encompasses a group of techniques which empower computers to learn tasks without explicit commands by drawing upon data to automatically discover meaningful patterns which can generate predictions at an individual level[8]. Previous studies have utilized ML to predict CRT outcomes using a variety of different features. Tokodi et al.[9] applied ML to predict all-cause mortality of CRT patients, and Feeny et al.[10] & Kalscheur et al.[11] applied ML to predict CRT response using clinical features and QRS morphology[12].

In this study, we aim to improve prediction of CRT response by accurately modeling the real-life multi-staged decision-making process of healthcare professionals using a multi-staged ML model guided by uncertainty quantification. This approach differs from previous studies as the results of all medical tests for all patients are not always necessary for prognostic predictions. Therefore, SPECT MPI acquisition can be omitted for patients whose data from the guidelines has low uncertainty for a decision.

## METHODS

### Patient population

This study draws from two multicenter clinical trials and a single center patient population for post hoc analysis. "Value of intraventricular synchronism assessment by gated-SPECT myocardial perfusion imaging in the management of HF patients submitted to cardiac resynchronization therapy" (IAEA VISION-CRT, n = 199)[13] is a non-randomized, multinational prospective cohort study. "SPECT Guided LV Lead Placement for Incremental Benefits to CRT Efficacy" (GUIDE-CRT, n = 137)[14] is a randomized, prospective cohort study. Additionally, patients from Taichung Veterans General Hospital were included (Taiwan-CRT, n = 31). After excluding patients lost-to-follow-up (n = 133), patients died before follow-up (n = 1), and patients with any missing features (ECG QRSd, PCI, MI, HTN, ACEI, or SPECT Concordance, n = 15), 218 patients remained for analysis. The endpoint was CRT response and was defined as an absolute increase in LVEF > 5% measured from the pre-implantation baseline to the $6\pm1$ month follow-up.

### Gated SPECT MPI acquisition and quantification

For VISION-CRT, the details of gated SPECT MPI acquisition can be found in our previous work[12,13]. For GUIDE-CRT, gated SPECT MPI was performed by SPECT systems with low-energy, general-purpose collimators within seven days before CRT implantation[14]. The gated SPECT scan was performed 60-90 minutes after injection with 25-30 mCi of 99mTc-MIBI at rest. Dual-head or triple-head camera systems were used for image acquisition following resting gated SPECT MPI protocol. Gated images were acquired with a photopeak window of the 99mTc set as a 20% energy window centered over 140 keV using the electrocardiographic R-R interval of 8 frames per cardiac cycle with a 50% beat acceptance window. All images were reconstructed using system-equipped iterative reconstruction algorithms with at least a 64×64 matrix and a 1.0 zoom factor. Emory Cardiac Toolbox (ECTb4, Atlanta, GA) was implemented to quantify short-axis and planar projection images for automated measurement of LV function and LV mechanical dyssynchrony (LVMD). For Taiwan-CRT, gated SPECT MPI was performed by dual-head SPECT/CT camera (Symbia T2, Siemens Healthcare)[15]. The gated SPECT scan was performed 60 minutes after injection with 20-30 mCi of 99mTc-MIBI at rest. Gated images were acquired using step-and-shoot acquisition with 25 seconds per stop and 32 stops over the 180° with a 64x64 matrix with 6.4 mm per pixel using the electrocardiographic R-R interval of 8 frames per cardiac cycle. Images were reconstructed using standard ordered subsets expectation maximization.

**Clinical Data**

Patient data included pre-CRT implantation baseline clinical features, ECG, and resting gated SPECT MPI variables. A total of 44 variables were extracted from these tests—baseline clinical features: age, gender, ethnicity, NYHA class, smoking history, presence of coronary artery disease (CAD), coronary artery bypass graft surgery (CABG), diabetes (DM), hypertension (HTN), myocardial infarction (MI), percutaneous coronary intervention (PCI), and current medical treatment including patient prescribed medication; ECG features: QRS duration and presence of LBBB; SPECT variables: myocardial mass, stroke volume, left ventricular scar percentage, LVEF, left ventricular end-systolic volume (ESV), end systolic eccentricity (ESE), end diastolic eccentricity (EDE), shape index (ESSI and EDSI), wall thickening (WT), summed rest score (SRS), both systolic and diastolic phase peak (PP), phase standard deviation (PSD), phase kurtosis (PK), phase skew (PS), and phase bandwidth (PBW). Additionally, the concordance between CRT LV lead position used and the optimal LV lead position identified by ECTb4 as the latest contracting viable site was recorded as a binary variable[16, 17].

**Uncertainty quantification**

Despite the impressive performance of ML across multiple domains, a significant hindrance to widespread implementation is the lack of uncertainty quantification causing a lack of transparency for clinicians[18]. Current methods in uncertainty quantification allow for interpretable measurements to monitor skeptical predictions. Kwon et al. applied uncertainty quantification via Bayesian neural networks for ischemic stroke lesion segmentation[19]. Linmans et al. utilized ensemble uncertainty quantification for tumor detection[20]. In the present study, pseudo-bootstrapped ensembles were employed. Ensembles are a technique which combines multiple base ML models to form a more informed prediction; using aggregation across these base model outputs allows for predictions as if the ensemble was a single standard ML model. Elastic-net logistic regression was selected as the base learner model for the ensembles. During training of the ensembles, two tunable hyperparameters were learned: (1) the fraction of training samples to use to train the models during random sampling and (2) the number of base models to use to form the ensemble. During inference, instead of predicting a single probability for CRT response, the ensemble provided multiple probabilities which depended on the number of base models. These probabilities were averaged to form an aggregated probability value, and the standard deviation was derived to estimate the uncertainty. It follows that a higher standard deviation indicates greater uncertainty of the sample, while a relatively lower standard deviation indicates less uncertainty of the sample[21,22].

**Multi-staged modeling**

In clinical practice, CRT follows a multi-staged decision-making process that can be generally separated into two stages: 1) an initial stage containing basic clinical variables and ECG features (stage 1); and 2) a second stage containing basic clinical variables, ECG features, and variables derived from gated SPECT MPI (stage 2). Considering the medical cost and associated time for the procedure, gated SPECT MPI could be considered for patients with high uncertainty at the first stage. We propose to use sequential trustworthy modeling via uncertainty quantification, to reduce clinical costs and improve efficiency by integrating basic clinical variables & ECG features with gated SPECT MPI features into a multi-staged model (Figure 1). Two key parameters are the standard deviation threshold and the midway threshold which control the tolerable level of uncertainty in a sample before SPECT acquisition becomes necessary. If the level of uncertainty is relatively small, the output from Ensemble 1 is used.

The overall data modeling pipeline is shown in Figure 2. A nested cross-validation structure was used where the outer training fold (train (9/10): n = 196, test (1/10): n = 22) was initially processed before two separate validation sets were sliced off (remaining train: n = 156, validation 1: n = 20, validation 2: n = 20). The data were pre-processed using centering/scaling and spatial sign, a technique to reduce the effect of outliers by projecting the data onto a multi-dimensional sphere. Feature selection was performed using Recursive Feature Elimination (RFE), a method of iteratively removing unimportant features to find the optimal set of features. From the original training fold, two validation sets were sliced to tune the multi-staged model's hyperparameters: standard deviation threshold and midway thresholds. These thresholds form the gated rule-based logic for determining if a patient needs to move from stage 1 into stage 2. The remaining training data were used in the inner cross-validation to tune the hyperparameters of the ensemble models, such as the number of base models in the ensemble, the percentage of samples to randomly select for each base model, and the respective base model's hyperparameters. These procedures are repeated twice for both stage 1 data (clinical and ECG features) and stage 2 data (clinical, ECG, and SPECT MPI features), thereby obtaining two ensemble models, ensemble 1 from stage 1 data and ensemble 2 from stage 2 data. After the two ensemble models were trained separately, the first validation set was used to tune the multi-stage model's hyperparameter thresholds; the metric for the hyperparameter optimization was a scaled weighted Area Under the (receiver operating characteristic) Curve (AUC). The original AUC of the model was constrained using a scaled weight function (Figure 3) to optimize a balance between predictions kept from Ensemble 1 and those cascading into Ensemble 2. First, the fraction of samples that did not require the stage 2 data were inputted into the scaling function which was then used to weight the AUC. Augmenting the AUC in this way constrains the multi-stage model to prevent only using Ensemble 2. Second, the hyperparameter tuning process was repeated for different scaling parameters in a range between [0.5, 9]. Afterwards, the second validation set was used to evaluate the learned thresholds from the different scaling parameters to select the best multi-staged model with the highest normal AUC. Finally, the best multi-stage model was evaluated using the outer test set.

For comparison to the multi-stage model, both Ensemble 1 and 2 performances were compared. Additionally, a new set of performance metrics were compared based on the 2012 ACCF/AHA/HRS guideline recommendations for CRT admission. CRT responders were predicted if the patient belonged to Class I or Class IIa CRT recommendations[23].

Additionally for robustness purposes, a simulation study is conducted to understand the effects of sample size on the performance of the model. The simulation involved random sampling with replacement of percentages of the training-fold within the 10-fold outer cross-validation; this fraction of training samples ranges from [0.1, 1.0] in increments of 0.10 resulting in 10 different samples that are then repeated three times. Ultimately, three times repeated cross validation is used. Afterwards, performance metrics of the model are aggregated to calculate means and standard deviations which show how the model performs under different perturbations to the training data size.

**Statistical analysis**

Performance metrics to evaluate the predictive ability of the different models included AUC, accuracy, sensitivity, and specificity. Feature importance was generated using permutation tests. For baseline analysis, categorical variables were expressed in counts and percentages with p-values generated via the chi-square test of independence while continuous variables were expressed as mean $\pm$ standard deviation with p-values generated via two-sample t-tests. Delong AUC test was used to generate confidence intervals and p-values. McNemar's test was used to compare sensitivity and specificity of the models for p-values.

## RESULTS

In total, 218 patients underwent CRT implantation which included an initial gated SPECT MPI and clinical assessment (Table 1). During a $6\pm1$ month follow-up, all the assessments were repeated. Overall, a total of 121 out of the 218 (55.5%) patients responded to CRT which was defined as an absolute increase in LVEF > 5% from baseline to follow-up.

Average performance metrics across the outer validation folds are presented in Table 2 for each of the models. For predicting CRT response, Ensemble 2 shows the highest AUC of 0.77 closely followed by the multi-stage model with 0.75 which has a slightly higher standard deviation, 0.08 vs. 0.10, respectively. Both these models outperformed Ensemble 1, with AUC of 0.70. The multi-stage model exhibited the highest accuracy of 0.71, outperforming both Ensemble 1 with 0.64 and Ensemble 2 with 0.69. Moreover, the accuracy standard deviation of the multi-stage model was the same as Ensemble 2, 0.11, but was outperformed by the Ensemble 1, 0.07. For sensitivity, Ensemble 2 displayed a performance of 0.72, slightly better than that of the multi-stage model, 0.70; however, the standard deviation was slightly higher, 0.14 vs. 0.13, respectively. Both models outperformed Ensemble 1 which had a sensitivity of 0.61. For specificity, the multi-stage model moderately outperformed both Ensembles 1 and 2 with values of 0.72, 0.67, and 0.65, respectively. However, the standard deviation for specificity of the multi-stage models, 0.20, was greater than that of Ensemble 1 and 2, 0.13 and 0.15, respectively. The corresponding AUC plots are shown in Figure 4.

The performance of the multi-stage model with respect to AUC and sensitivity was only slightly inferior to Ensemble 2 but outperformed in terms of accuracy and specificity. Of note, the multi-stage model *only required 52.7% (25$^{th}$ percentile: 5.7%, 50$^{th}$ percentile: 50.0%, 75$^{th}$ percentile: 77.3%) of the patients to acquire the additional SPECT MPI modality*. Moreover, when compared to the guidelines model, the multi-stage model shows impressive accuracy, 0.71 vs. 0.53, respectively. The guideline performance is higher in terms of sensitivity, 0.75 vs. 0.70, when compared to the multi-stage model but this is outweighed by the sizeable boost in specificity by the multi-stage model, 0.72 vs. 0.26, respectively. All models generated trended towards improvement over the guideline model. Statistical testing on AUC, sensitivity, and specificity is presented in Table 4. While no significance was found between the multi-stage mode, Ensemble 1, and Ensemble 2, all models had significant differences in specificity compared to the Guideline model with p-values <0.001, <0.001, and <0.001, respectively.

The importance of each variable for Ensembles 1 and 2 is presented in Figure 5. The feature importance was averaged within the ensemble of each fold, then the absolute value was aggregated across folds. Ensemble 1 makes heavy use of race parameters, age, CAD, and NYHA class 3. Ensemble 2 heavily uses LVEF, race parameters, ESSI, ESE, and EDV. The most important variable in Ensemble 2 was LVEF. Hyperparameters for the multi-stage model are presented in table 3.

Additionally, Figure 6 displays the results of a simulation study conducted to understand the effects of sample size on the performance of the model The colored lines display the average performance, while the ribbon displays the standard deviation. The AUC has a sharp increase after increasing the percent of training samples beyond 20%, but then plateaus with a slightly decreasing amount of variation as the fraction of training data increases. This trend is similar for accuracy and specificity, while sensitivity increases moderately until 40% of the training data is used and the variation decreases correspondingly. The plateau of performance from the sample size indicates that the model is receiving decreasing benefits as the training sample size increases. Therefore, it is likely that increasing the sample size will not drastically improve performance; however, it should be noted that the test sets within the cross-validation folds are kept unchanged. Hence, developing new data features is paramount to increasing the model capabilities for predicting CRT response rather than sample size which also remains important.

## DISCUSSION

In this study, we developed a multi-staged based ML ensemble model which utilized uncertainty quantification and internal logic to predict CRT response using different stages of data. The data for this study was constructed from 218 patients from the VISION-CRT and GUIDE-CRT clinical trials. The multi-stage model's performance, which includes the internal logic rules for cascading patients to the next stage was compared with the performance of two composite ensemble models. Ensemble 1 contained only baseline clinical and ECG variables, and Ensemble 2 contained the first stage's data in addition to the features derived from gated SPECT MPI. The multi-stage model always utilized Ensemble 1 and only used Ensemble 2 if a given patient's CRT response prediction's uncertainty was below the uncertainty threshold or midway threshold. The Ensemble 2 model achieved the highest AUC and sensitivity, while the multi-stage model achieved the highest accuracy and specificity. Considering that the Ensemble 2 model always had the additional gated SPECT MPI modality while the multi-stage model only used this modality on average 52.7 % of the time, shows impressive potential. Although, these models still show moderate accuracy, utilizing higher optimized and non-linear models has the potential for improved performance.

Many studies have applied ML to predict CRT response[9,10,11,12]; however, these studies have implicitly assumed that all modalities of data (biological variables, baseline characteristics, ECG, and gated SPECT MPI) will be available for the ML models at inference. In contrast the multi-stage model is able to form patient predictions in a more flexible manor which takes into account the cost of acquiring new data; moreover, through this modeling procedure, not only can we generate feature importance like standard ML algorithms, the frequency of data use at each stage can be calculated providing more nuanced information regarding the overall importance of different tests towards the ultimate goal.

Both mechanical and electrical dyssynchrony influence the outcome of CRT response; however, they are not often found concurrently in a given patient[24]. The creation of CRT arose to remedy mechanical dyssynchrony with the end goal of improving cardiac function[24,25,26]. However, when generating the patient selection criteria for CRT, since electrical and mechanical dyssynchrony are correlated in addition to the relative ease of measuring electrical dyssynchrony over mechanical dyssynchrony, it was presumed that only electrical dyssynchrony would be necessary. While QRS duration has proven important for predicting CRT response, in many specific instances it has shown otherwise[27]. Moreover, continual research into CRT response has highlighted that electrical and mechanical dyssynchrony are not equivalent nor interchangeable[28,29]. Historically, this false equivalence partially can explain the high rate of CRT non-response[30]. Nevertheless, our experiments reveal that while electrical dyssynchrony is not perfect, in many cases it is sufficient to indicate the need for CRT; furthermore, information pertaining to mechanical dyssynchrony is not always imperative when predicting CRT response. Hence, we showed that a multi-stage model that does not always require gated SPECT MPI data can perform about as well as a model which always utilizes SPECT MPI. Interestingly, neither LBBB nor QRS duration made it to the top 10 important variables in the Ensemble 2 or even the Ensemble 1 which further emphasizes the need for rigorous clinical investigation into the power of electrical dyssynchrony.

CRT response is a difficult task to predict due to the complex origins of HF[31]. However, by modeling the prediction task as a sequential diagnostic process, the need for additional patient data (i.e. additional medical tests) can be incorporated into the ML models via gated rule-based logic. Using uncertainty quantification, further data can be acquired when necessary. Moreover, the associated cost and wasted time of additional tests can be circumvented for HF patients.

The generalizability and representativeness of the multi-stage was strengthened by the data sources: the VISION-CRT and GUIDE-CRT multicenter, multinational trials. Spanning 10 centers across 8 different countries including Brazil, Chile, Columbia, Cuba, Indian, Mexico, Pakistan, and Spain, the VISION-CRT trial provided a significant spread of HF patients worldwide. Moreover, GUIDE-CRT encompassed 19 centers across China.

**Limitations and Future Work**

The analysis from this study stemmed from medical centers that used *Emory Cardiac Toolbox* to process and generate derived variables from gated SPECT MPI. It is known that different software tools for processing SPECT data have the potential to generate different results when calculating LV function and LVMD[32,33,34]. Therefore, comparing results from other available commercial software packages may yield different results. Moreover, the different studies used different acquisition protocols which can influence cardiac function metrics, such as phase SD and bandwidth. For future work, encoding the different acquisition protocol as a predictor itself may prove useful in modeling, specifically with differences in SPECT MPI derived parameters. The small number of patients (218) enrolled in this study also has the potential to increase the variability of the analysis.

In addition to clinical and ECG parameters, it is common to use features derived from echocardiography, specifically, speckle-tracking variables (STE). For future work, incorporating STE features as a new stage in the CRT decision-making process, as a replacement for gated

SPECT MPI, or as an insertion between the two existing stages may yield interesting results. This comparative analysis could not be conducted in this study as STE data was not collected in VISION-CRT.

## Conclusions

The proposed multi-staged method using ML to predict CRT response combining multi-staged uncertainty quantifying ensembles represents a step towards accurately modeling the complex decision-making process faced by clinicians. The results of this approach highlight the great potential of modeling tasks in a sequential fashion which provides more informative clinical interpretation for patient care.

## Declarations

### *Ethics Approval and Consent to Participate*

The studies were approved by an institutional review committee and all subjects gave informed consent.

### *Acknowledgements*

The authors would like to thank the International Atomic Energy Agency (IAEA) for providing access to the data of the multicenter trial: “Value of intraventricular synchronism assessment by gated-SPECT myocardial perfusion imaging in the management of heart failure patients submitted to cardiac resynchronization therapy” (IAEA VISION-CRT), Coordinated Research Protocol E1.30.34. The authors would also like to thank the Department of Cardiology, the First Affiliated Hospital Nanjing Medical University for providing access to the data of the multicenter trial: “SPECT Guided LV Lead Placement for Incremental Benefits to CRT Efficacy” (GUIDE-CRT).

### *Funding*

This research was supported by a Michigan Technological University Research Excellence Fund Research Seed grant (PI: Weihua Zhou), a research seed grant from Michigan Technological University Health Research Institute (PI: Weihua Zhou), a grant from the National Institutes of Health, USA (1R15HL172198) (PI: Weihua Zhou), and a Michigan Technological University Health Research Institute Fellowship (PI: Kristoffer Larsen). This work was also supported by the National Natural Science Foundation of China (82070521), the Clinical Competence Improvement

Project of Jiangsu Province Hospital (JSPH-MA-2020-3), and the Project on New Technology of Jiangsu Province (JX233C202103).

***Conflict of Interest Disclosure Statement***

All authors declare that there are no conflicts of interest.

***Data Availability Statement***

The patient data that supports the findings of this study is available from the corresponding author, Weihua Zhou, upon reasonable request. The computer source code for this study is available at https://github.com/MIILab-MTU/CRT_MultiStageML_Uncertainty.

**REFERENCES**

[1] Virani SS, Alonso A, Benjamin EJ, *et al.* Heart disease and stroke statistics—2020 update: A report from the American Heart Association. *Am Heart Assoc* 2020; **141**: 139–596.

[2] Urbich M, Globe G, Pantiri K, Heisen M, Bennison C, Wirtz HS, Di Tanna GL. A systematic review of medical costs associated with heart failure in the USA. *Pharmacoeconomics* 2020; **38**: 1219-1236.

[3] Auricchio A, Stellbrink C, Sack S, *et al.* Long-term clinical effect of hemodynamically optimized cardiac resynchronization therapy in patients with heart failure and ventricular conduction delay. *J Am Coll Cardiol* 2002; **39**: 2026-2033.

[4] Auricchio A, Prinzen FW. Non-responders to cardiac resynchronization therapy: the magnitude of the problem and the issues. *Circ J*. 2011; **7**: 521–7.

[5] Donal E, Hubert A, Le Rolle V, *et al*. New Multiparametric Analysis of Cardiac Dyssynchrony: Machine Learning and Prediction of Response to CRT. *JACC Cardiovasc Imaging* 2019; **12**:1887-1888.

[6] Zhou W, Garcia EV. Nuclear image-guided approaches for CRT. Current Cardiology Reports 2016; **18**: 1-11.

[7] He Z, Garcia EV, Zhou W. Chapter 25 – Nuclear Image-Guided Methods for Cardiac Resynchronization Therapy. Nuclear imaging guiding cardiac resynchronization therapy in Nuclear Cardiology: Basic and Advanced Concepts in Clinical Practice. Springer-Nature. Book chapter. 2021.

[8] Janiesch C, Zschech P, Heinrich K. Machine learning and DL. *Electron Markets* 2021; **31**: 685–695.

[9] Tokodi M, Richard SW, Kovács A, *et al.* Machine learning-based mortality prediction of patients undergoing cardiac resynchronization therapy: the SEMMELWEIS-CRT score. *Eur Heart J* 2020; **41**: 1747–1756.

[10] Feeny AK, Rickard J, Patel D, *et al.* Machine Learning Prediction of Response to Cardiac Resynchronization Therapy: Improvement Versus Current Guidelines. *Circ Arrhythm Electrophysiol 2019*; **12**; e007316.

[11] Kalscheur MM, Kipp RT, Tattersall MC, *et al*. Machine Learning Algorithm Predicts Cardiac Resynchronization Therapy Outcomes: Lessons From the COMPANION Trial. *Circ Arrhythm Electrophysiol* 2018; **11:** e005499.

[12] de A Fernandes F, Larsen K, He Z, *et al.* A machine learning method integrating ECG and gated SPECT for cardiac resynchronization therapy decision support. *Eur J Nucl Med Mol Imaging* 2023; **50**: 3022-3033.

[13] Peix A, Karthikeyan G, Massardo T, *et al.* Value of intraventricular dyssynchrony assessment by gated-SPECT myocardial perfusion imaging in the management of heart failure patients undergoing cardiac resynchronization therapy (VISION-CRT). *J Nucl Cardiol* 2021; **28**: 55-64.

[14] Zou J, Hua W, Su Y, *et al.* SPECT-Guided LV Lead Placement for Incremental CRT Efficacy: Validated by a Prospective, Randomized, Controlled Study. *JACC Cardiovasc Imaging* 2019; **12**: 2580-2583.

[15] Hung GU, Zou J, He Z, *et al.* Left-ventricular dyssynchrony in viable myocardium by myocardial perfusion SPECT is predictive of mechanical response to CRT. *Ann Nucl Med* 2021; **35**: 947-954.

[16] Boogers MJ, Chen J, Van Bommel RJ, *et al.* Optimal left ventricular lead position assessed with phase analysis on gated myocardial perfusion SPECT. *Eur J Nucl Med Mol Imaging* 2011; **38**: 230–238.

[17] Zhou W, Tao N, Hou X, *et al.* Development and validation of an automatic method to detect the latest contracting viable left ventricular segments to assist guide CRT therapy from gated SPECT myocardial perfusion imaging. *J Nucl Cardiol* 2017; **25**: 1948-1957.

[18] Begoli E, Bhattacharya T, Kusnezov D. The need for uncertainty quantification in machine-assisted medical decision making. *Nat Mach Intell 2019*; **1**: 20-23.
[19] Kown Y, Won JH, Kim BJ. Uncertainty quantification using Bayesian neural networks in classification: Application to biomedical image segmentation. *Comput Stat Data Anal* 2019; **142**: e106816.
[20] Heid E, McGill C, Vermeire F, Green W. Characterizing Uncertainty in Machine Learning for Chemistry. Am Chem Soc *2023*; **63**: 4012-4029.
[21] Jacobs R, Morgan D. Opportunities and Challenges for Machine Learning in Material Science. Annu Rev Mater Res *2020*; **50**: 71-103.
[22] Linmans J, Elfwing S, van der Laak J, Litjens G. Predictive uncertainty estimation for out-of-distribution detection in digital pathology. Med Image Anal 2023; **83**: e102655.
[23] Tracy CM, Epstein AE, Darbar D, et al. 2012 ACCF/AHA/HRS focused update of the 2008 guidelines for device-based therapy of cardiac rhythm abnormalities: a report of the American College of Cardiology Foundation/American Heart Association Task Force on Practice Guidelines and the Heart Rhythm S. Circulation 2012; **126**: 1784–1800.
[24] Fudim M, Borges-Neto S. A troubled marriage: When electrical and mechanical dyssynchrony don't go along. *J Nucl Cardiol* 2019; **26**: 1240-1242.
[25] Lozano I, Bocchiardo M, Achtelik M, *et al*. Impact of biventricular pacing on mortality in a randomized crossover study of patients with heart failure and ventricular arrhythmias. *Pacing Clin Electrophysiol* 2000; **23**:1711-1712.
[26] Auricchio A, Stellbrink C, Sack S, *et al*. Long-term clinical effect of hemodynamically optimized cardiac resynchronization therapy in patients with heart failure and ventricular conduction delay. *J Am Coll Cardiol* 2002; **39**:2026-2033.
[27] Yu CM, Zhang Q, Chan YS, Chan CK, Yip GW, Kum LC, Wu EB, Lee PW, Lam YY, Chan S, Fung JW. Tissue Doppler velocity is superior to displacement and strain mapping in predicting left ventricular reverse remodeling response after cardiac resynchronization therapy. *Heart* 2006; **92**:1452-1456.
[28] Abraham T, Kass D, Tonti G, Tomassoni GF, Abraham WT, Bax JJ, et al. Imaging cardiac resynchronization therapy. *J Am Coll Cardiol Cardiovas Img* 2009; **2**:486–497.

[29] AlJaroudi W, Chen J, Jaber WA, Lloyd SG, Cerqueira MD, Marwick T. Nonechocardiographic imaging in evaluation for cardiac resynchronization therapy. *Circulation Cardiovasc Img* 2011; **4**:334–43.

[30] Carita P, Corrado E, Pontone G, *et al*. Non-responders to cardiac resynchronization therapy: Insights from multimodality imaging and electrocardiography. A brief review. *Int J Cardiol* 2016; **225**:402–407.

[31] Naqvi SY, Jawaid A, Goldenberg I, Kutyifa V. Non-response to Cardiac Resynchronization Therapy. *Curr Heart Fail Rep* 2018; **15**: 315-321.

[32] Gumuser G, Parlak Y, Topal G, Batok D, Ruksen E, Bilgin E. Comparison between ECTb and QGS for assessment of left ventricular function. *J Nucl Med* 2007; **48**; 408.

[33] Nichols K, Santana CA, Folks R, *et al.* Comparison between ECTb and QGS for assessment of left ventricular function from gated myocardial perfusion SPECT. *J Nucl Cardiol* 2002; **9**: 285-293.

[34] Okuda K, Nakajima K, Matsuo S, *et al.* Comparison of diagnostic performance of four software packages for phase dyssynchrony analysis in gated myocardial perfusion SPECT. *Eur J Nucl Med Mol Imaging* 2017; **7**: 1-9.

Figure 1. Multi-staged decision-making process.

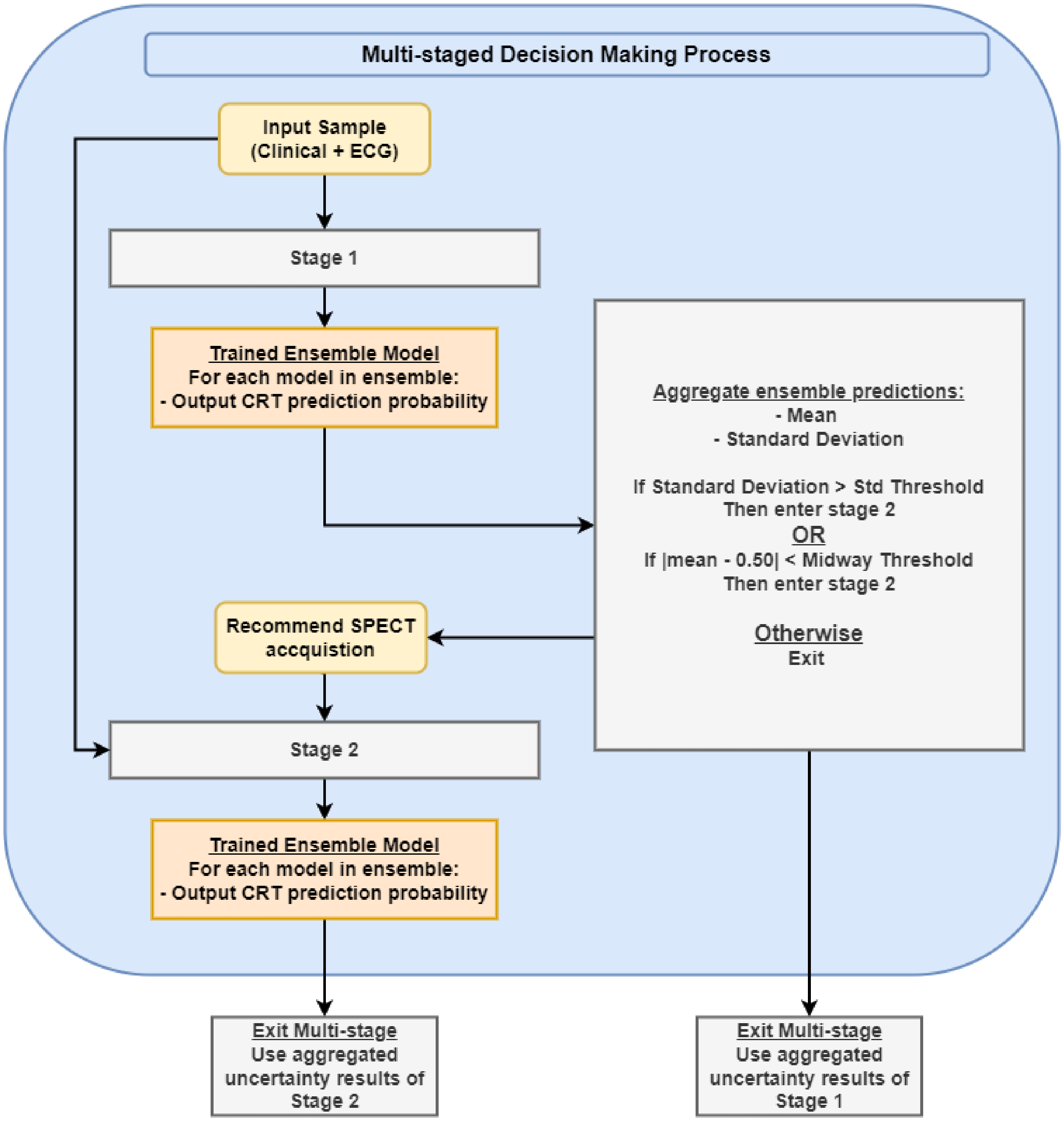

Figure 2. Data modeling pipeline.

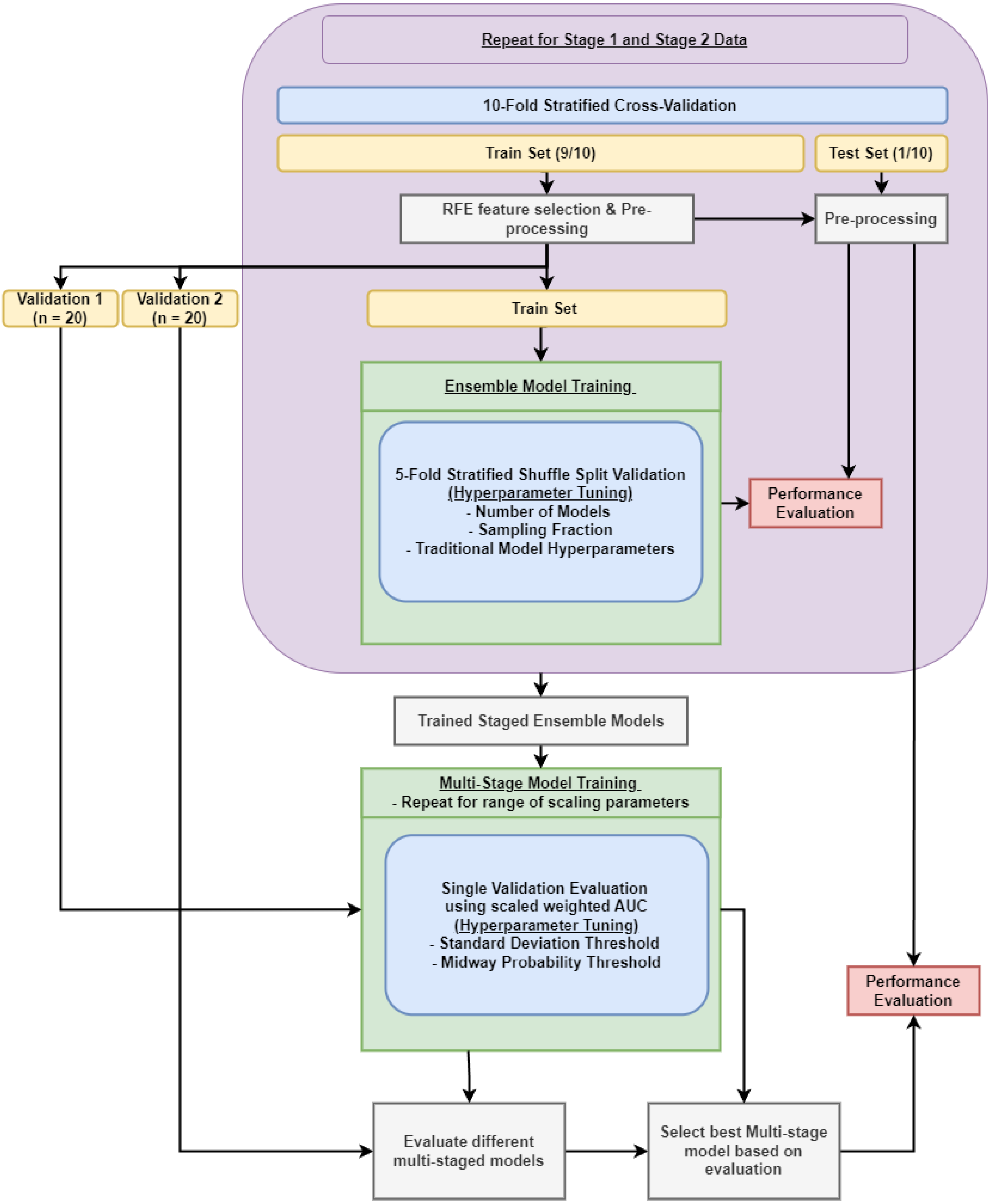

Figure 3. Scaled weighted AUC function.

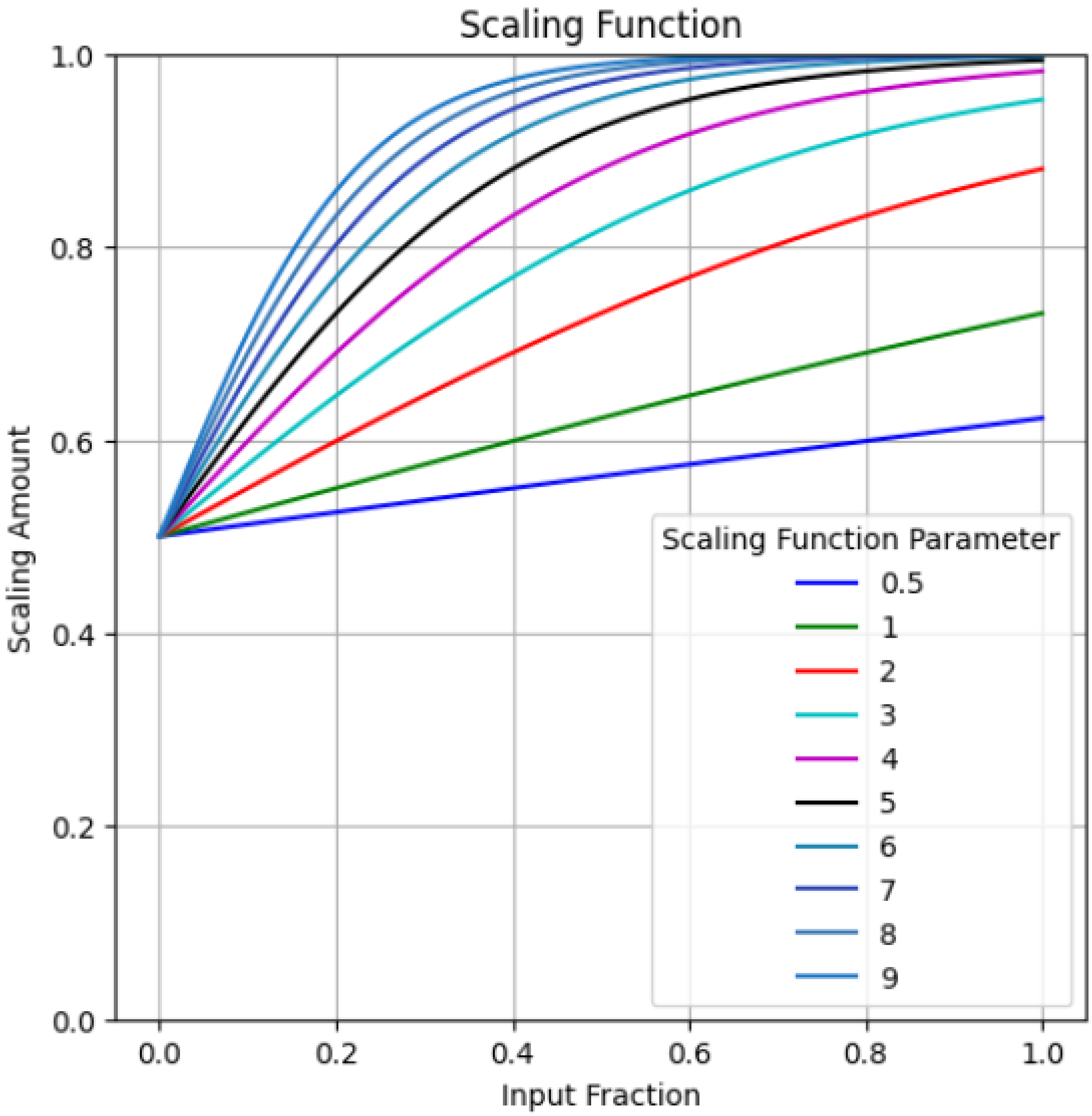

$$f(frac, param) = 1 - \frac{1}{e^{frac \cdot param} + 1}$$

Figure 4. AUC plots for ensemble models and multi-stage model.

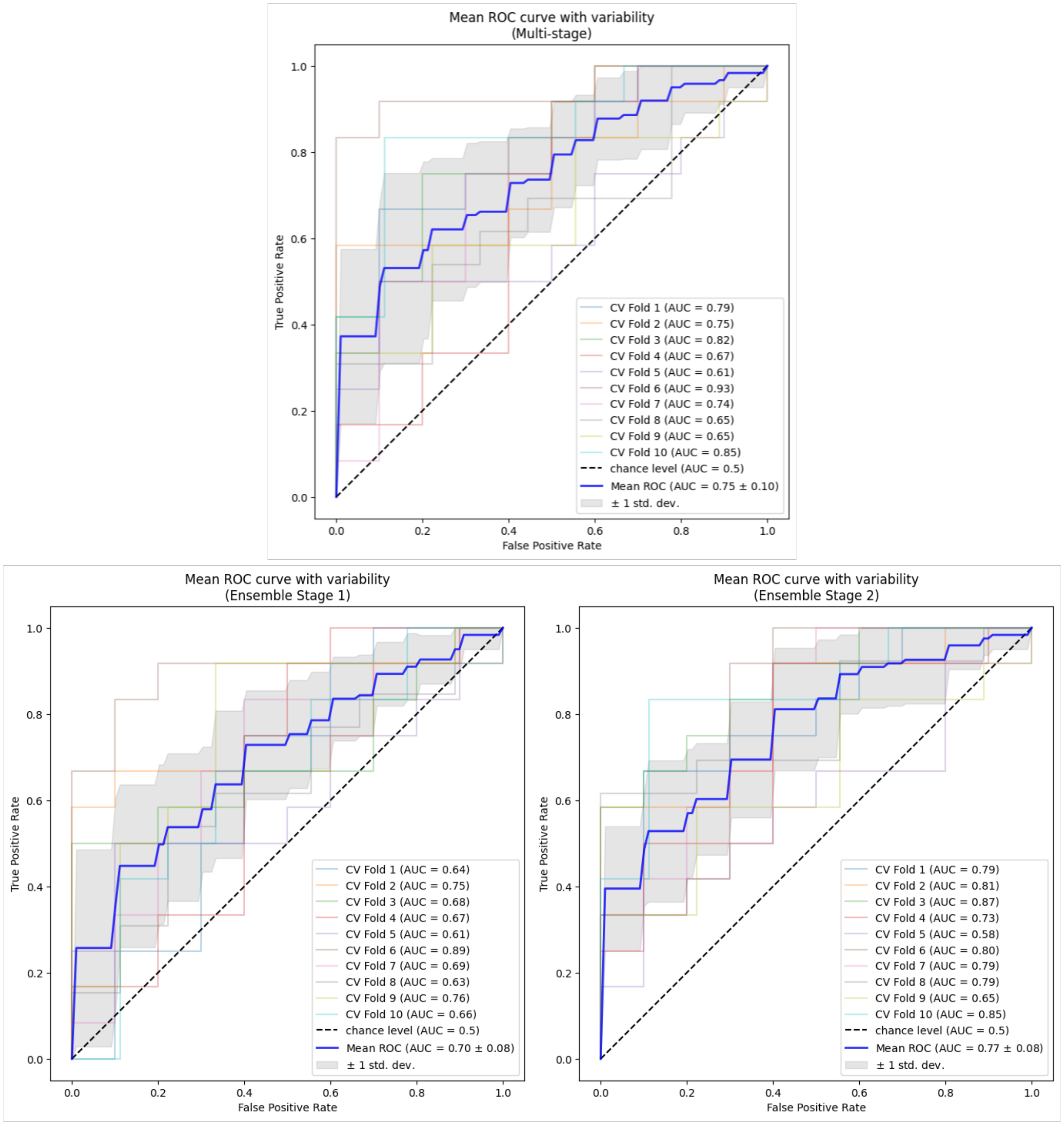

Figure 5. Feature importance of stage 1 and stage 2 ensemble models.

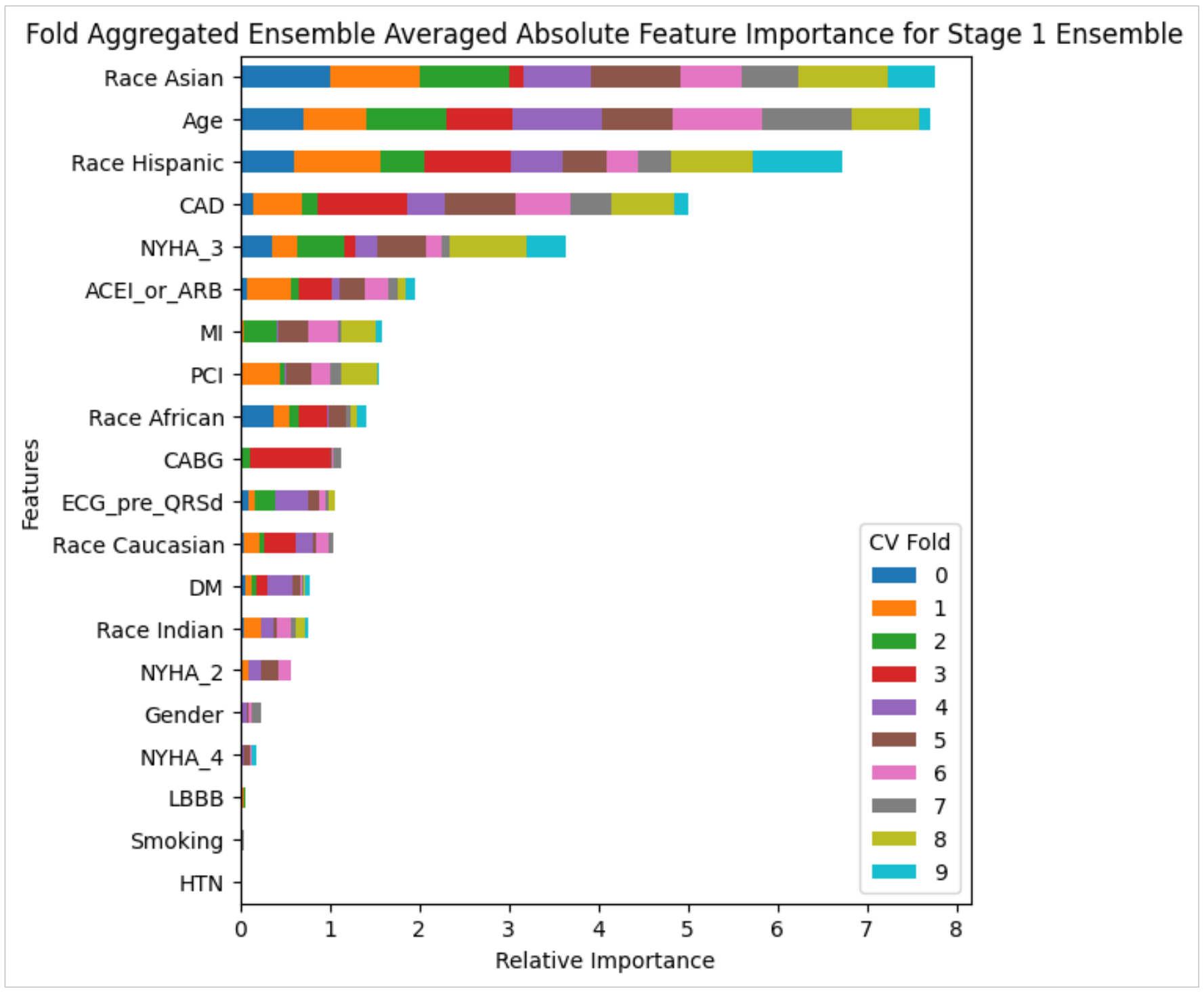

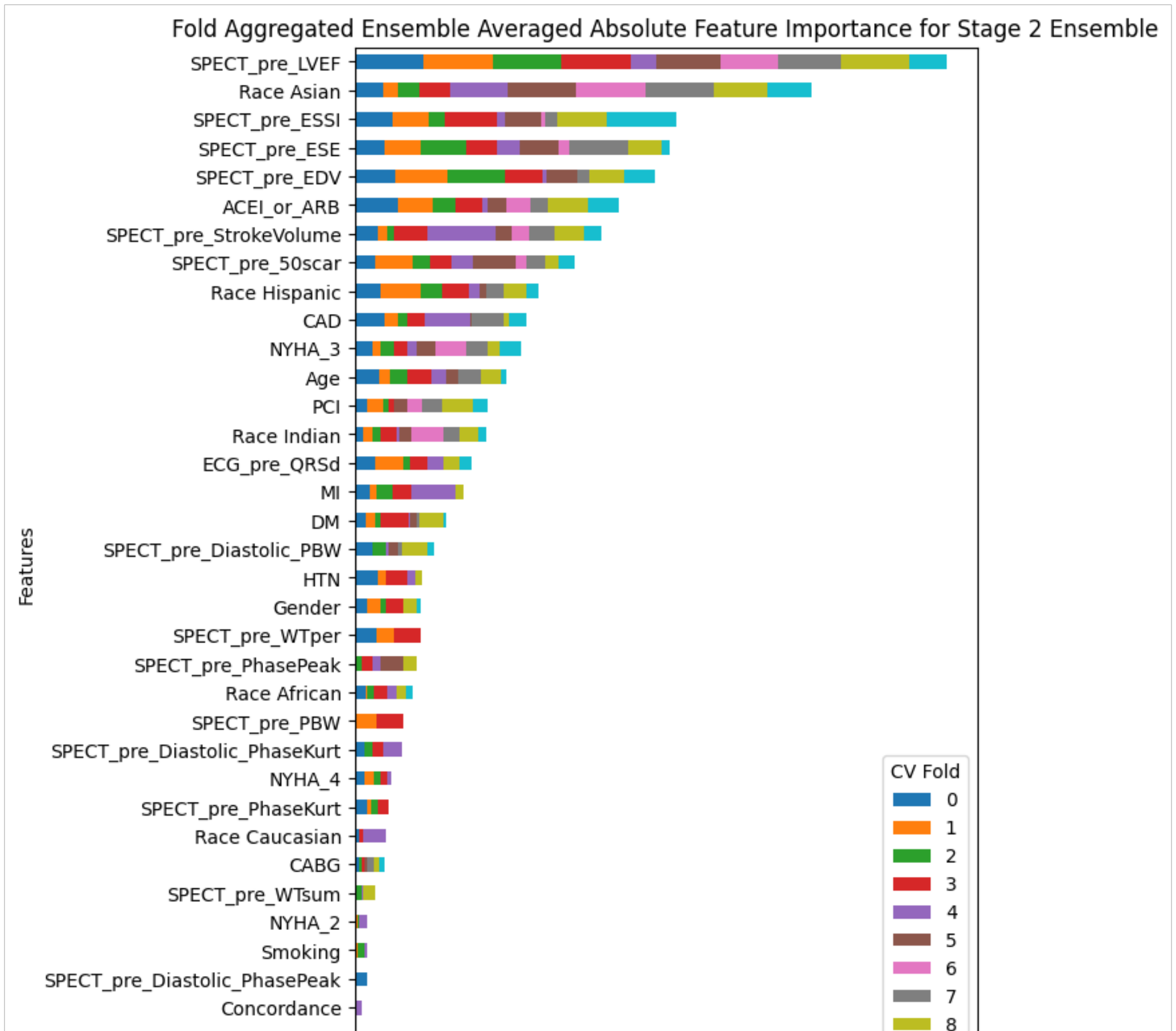

Figure 6. Sample size simulations for multi-stage model (mean with standard deviation).

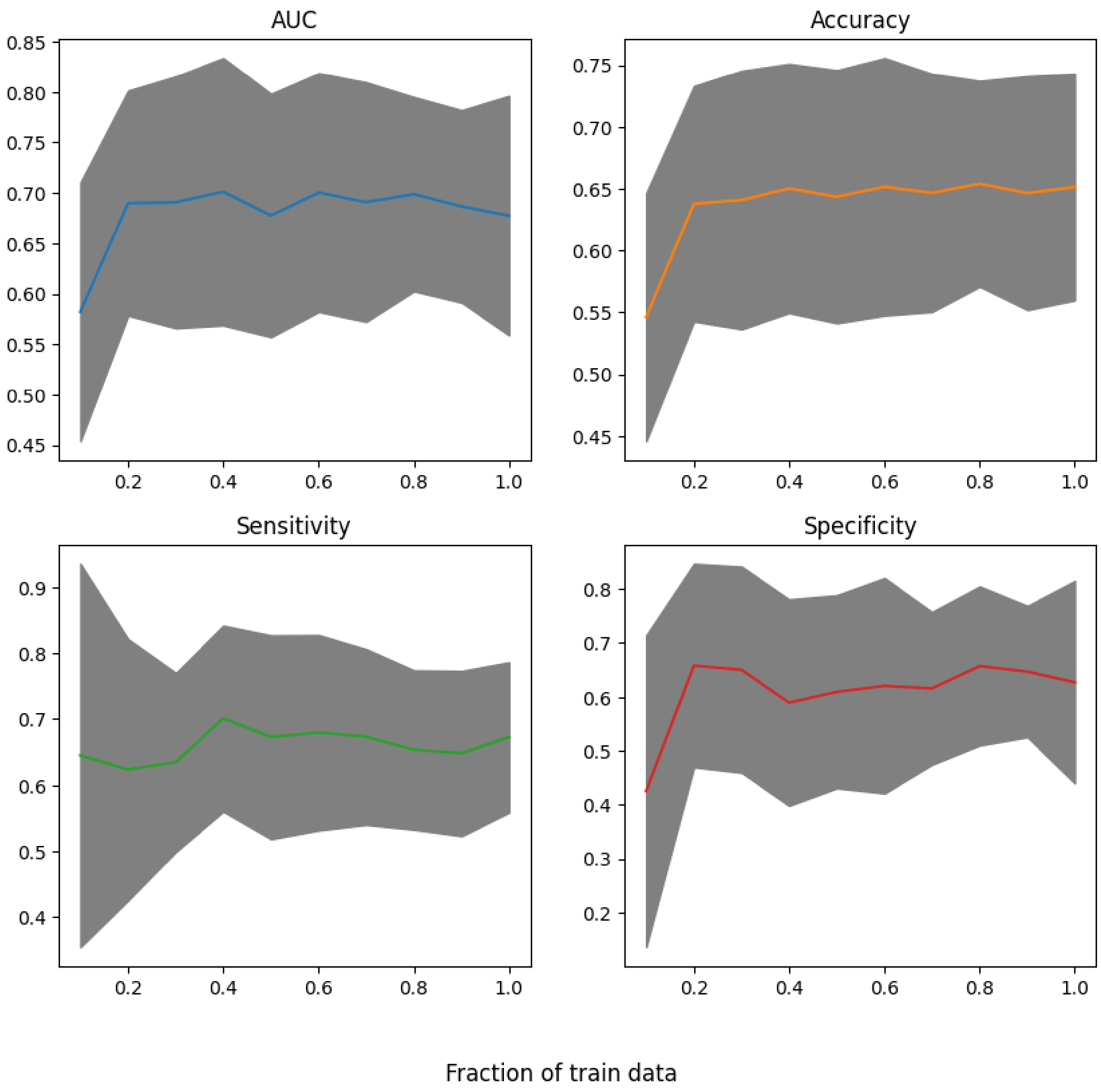

Table 1. Baseline characteristics of the enrolled patients.

| Variable | Overall (n=218) | Response (n=121, 55.5%) | Non-response (n=97, 44.5%) | P-value |
|---|---|---|---|---|
| **Age (yrs.)** | 62.0 ± 11.8 | 60.5 ± 12.1 | 63.2 ± 11.5 | 0.098 |
| **Male** | 133 (61.0%) | 63 (64.9%) | 70 (57.9%) | 0.353 |
| **Race** | | | | |
| **African** | 15 (6.9%) | 9 (9.3%) | 6 (5.0%) | 0.326 |
| **Asian** | 76 (34.9) | 21 (21.6%) | 55 (45.5%) | <.0001* |
| **Caucasian** | 21 (9.6%) | 11 (11.3%) | 10 (8.3%) | 0.593 |
| **Hispanic** | 76 (34.9%) | 45 (46.4%) | 31 (25.6%) | 0.002* |
| **Indian** | 30 (13.7%) | 11 (11.3%) | 19 (15.7%) | 0.465 |
| **Smoking** | 41 (18.8%) | 18(18.6%) | 23 (19.0%) | 1.0 |
| **DM** | 53 (24.3%) | 27 (27.8%) | 26 (21.5%) | 0.354 |
| **HTN** | 117 (53.7%) | 57 (58.8%) | 60 (49.6%) | 0.225 |
| **MI** | 34 (15.6%) | 23 (23.7%) | 11 (9.1%) | 0.006* |
| **CAD** | 68 (31.2%) | 41 (42.3%) | 27 (22.3%) | 0.003* |
| **CABG** | 3 (1.4%) | 1 (1.0%) | 2 (1.7%) | 1.0 |
| **PCI** | 12 (5.5%) | 7 (7.2%) | 5 (4.1%) | 0.488 |
| **NYHA** | | | | |
| **II** | 59 (27.1%) | 21 (21.6%) | 38 (31.4%) | 0.145 |
| **III** | 127 (58.3%) | 65 (67.0%) | 62 (51.2%) | 0.027* |
| **IV** | 32 (14.7%) | 11 (11.3%) | 21 (17.4%) | 0.292 |
| **ACEI/ARB** | 179 (82.1%) | 73 (75.3%) | 106 (87.6%) | 0.029* |
| **ECG** | | | | |
| **QRSd** | 158.6 ± 27.2 | 157.8 ± 27.3 | 159.2 ± 27.1 | 0.718 |
| **LBBB** | 214 (98.17%) | 96 (99.0%) | 118 (97.5%) | 0.776 |
| **SPECT** | | | | |
| **SRS** | 18.2 ± 12.2 | 21.6 ± 12.5 | 15.5 ± 11.3 | <.0001* |
| **ESV** | 192.6 ± 108.0 | 207.1 ± 118.7 | 181.0 ± 97.5 | 0.082 |
| **LVEF** | 27.7 ± 11.0 | 28.8 ± 11.1 | 26.9 ± 10.9 | 0.207 |
| **Mass** | 215.2 ± 58.4 | 224.3 ± 61.6 | 207.9 ± 54.9 | 0.042* |
| **Stroke Volume** | 62.9 ± 23.2 | 71.0 ± 24.1 | 56.4 ± 20.1 | <.0001* |
| **WT %** | 22.0 ± 16.9 | 24.9 ± 16.6 | 19.8 ± 16.9 | 0.026* |
| **WT Sum** | 11.2 ± 8.7 | 12.7 ± 8.5 | 10.1 ± 8.6 | 0.027* |
| **Concordance** | 49 (22.5%) | 25 (25.8%) | 24 (19.8%) | 0.379 |
| **Scar %** | 22.7 ± 14.0 | 27.0 ± 16.0 | 19.2 ± 11.1 | <.0001* |
| **Diastolic** | | | | |
| **PBW** | 169.4 ± 80.5 | 171.6 ± 78.9 | 167.7 ± 82.0 | 0.725 |
| **PK** | 8.4 ± 7.3 | 8.8 ± 7.8 | 8.1 ± 6.8 | 0.456 |
| **PS** | 2.5 ± 0.8 | 2.5 ± 0.9 | 2.5 ± 0.8 | 0.819 |
| **PP** | 221.0 ± 0.8 | 220.6 ± 34.2 | 221.4 ± 43.7 | 0.886 |
| **PSD** | 52.3 ± 20.7 | 52.7 ± 20.0 | 52.0± 21.4 | 0.803 |
| **Systolic** | | | | |
| **PBW** | 158.3 ± 77.5 | 159.8 ± 77.3 | 157.1 ± 77.9 | 0.798 |
| **PK** | 132.5 ± 8.1 | 9.2 ± 9.3 | 7.5 ± 6.9 | 0.134 |
| **PP** | 132.5 ± 37.2 | 131.0 ± 32.8 | 133.6 ± 40.4 | 0.599 |
| **PSD** | 50.0 ± 20.6 | 50.5 ± 20.8 | 49.7 ± 20.5 | 0.758 |
| **EDE** | 0.6 ± 0.2 | 0.5 ± 0.2 | 0.6 ± 0.1 | 0.002* |
| **EDSI** | 0.8 ± 0.1 | 0.8 ± 0.1 | 0.7 ± 0.1 | 0.003* |
| **EDV** | 255.5 ± 118.0 | 278.2 ± 129.4 | 237.4 ± 105.1 | 0.013* |
| **ESE** | 0.6 ± 0.2 | 0.6 ± 0.2 | 0.6 ± 0.1 | <.0001* |
| **ESSI** | 0.8 ± 0.1 | 0.8 ± 0.1 | 0.8 ± 0.1 | 0.001* |

Data are expressed as mean ± standard deviation or count (percentage)

* P-value < 0.05

DM, Diabetes Mellitus; HTN, Hypertension; MI, Myocardial Infarction; CAD, Coronary Artery Disease; CABG, Coronary Artery Bypass Graft surgery; PCI, Percutaneous Coronary Intervention; NYHA, New York Heart Association; ACEI/ARB, Angiotensin-Converting-Enzyme Inhibitors/Angiotensin Receptor Blockers; SRS, Summed Rest Score; ESV, End-Systolic Volume; WT, wall thickening; Scar %, percentage non-viable LV; PBW, Phase Bandwidth; PK, Phase Kurtosis; PS, Phase Skew; PP, Phase Peak; PSD, Phase Standard Deviation; EDE, End-Diastolic Eccentricity; EDSI, End-Diastolic Shape Index; EDV, End-Diastolic Volume; ESE, End-Systolic Eccentricity; ESSI, End-Systolic Shape Index.

Table 2. Performance of ensemble models and multi-stage model.

| Model | AUC | Accuracy | Sensitivity | Specificity |
|---|---|---|---|---|
| Multi-stage | 0.75 (0.10) | **0.71** (0.11) | 0.70 (0.13) | **0.72** (0.20) |
| Ensemble 1 | 0.70 (**0.08**) | 0.64 (**0.07**) | 0.61 (**0.09**) | 0.67 (**0.13**) |
| Ensemble 2 | **0.77** (**0.08**) | 0.69 (0.11) | **0.72** (0.14) | 0.65 (0.15) |
| Guideline | N/A | 0.53 | 0.75 | 0.26 |

Performance metrics are presented as mean (standard deviation). **Bold** represents best performance values.

Table 3. Multi-stage model hyperparameters.

| Outer CV Fold | Std Threshold | Midway Threshold | Scaling Weight | Ensemble 1 | Ensemble 2 |
|---|---|---|---|---|---|
| 0 | 0.01 | 0.10 | 0.5 | 25 (0.95) | 37 (0.70) |
| 1 | 0.15 | 0.04 | 8 | 25 (0.95) | 34 (0.95) |
| 2 | 0.06 | 0.06 | 9 | 25 (0.70) | 37 (0.70) |
| 3 | 0.20 | 0.02 | 0.5 | 34 (0.70) | 25 (0.70) |
| 4 | 0.19 | 0.02 | 1 | 25 (0.95) | 25 (0.95) |
| 5 | 0.20 | 0.03 | 0.5 | 28 (0.84) | 40 (0.95) |
| 6 | 0.15 | 0.06 | 0.5 | 49 (0.71) | 40 (0.95) |
| 7 | 0.20 | 0.02 | 0.5 | 25 (0.70) | 34 (0.85) |
| 8 | 0.01 | 0.10 | 0.5 | 25 (0.86) | 49 (0.86) |
| 9 | 0.01 | 0.10 | 0.5 | 25 (0.95) | 37 (0.85) |

Hyperparameters for the ensemble models are presented as: number of models (fraction of samples randomly selected for each individual model).

Table 4. Statistical testing p-values for different metrics.

| | Multi-stage | Ensemble 1 | Ensemble 2 | Guideline |
|---|---|---|---|---|
| Multi-stage | | | | |
| Ensemble 1 | AUC: 0.104<br>Sens: 0.180<br>Spec: 0.096 | | | |
| Ensemble 2 | AUC: 0.357<br>Sens: 0.564<br>Spec: 0.096 | AUC: 0.105<br>Sens: 0.225<br>Spec: 1.000 | | |
| Guideline | AUC: N/A<br>Sens: 0.670<br>Spec: **<0.001** | AUC: N/A<br>Sens: 0.336<br>Spec: **<0.001** | AUC: N/A<br>Sens: 1.000<br>Spec: **<0.001** | |

The p-values follow the order: Delong AUC, McNemar's sensitivity, McNemar's specificity.